\documentclass[journal]{IEEEtran}

\ifCLASSINFOpdf
\else
   \usepackage[dvips]{graphicx}
\fi
\usepackage{url}

\usepackage{graphicx}

\usepackage{cite}
\usepackage{amsmath}
\usepackage{multirow}
\usepackage{multicol}
\usepackage{subfigure}
\usepackage{float}
\usepackage[dvipsnames]{xcolor}

\begin{document}

\title{RotationDrag: Point-based Image Editing with Rotated Diffusion Features}

\author{Minxing Luo, Wentao Cheng, and Jian Yang
\thanks{Corresponding authors: Wentao Cheng and Jian Yang.}
\thanks{Minxing Luo, Wentao Cheng and Jian Yang are with VCIP, CS, the College of Computer Science, Nankai University, Tianjin, 300350, China (e-mail: lmx@mail.nankai.edu.cn; wentaocheng@nankai.edu.cn; csjyang@nankai.edu.cn)}
}

\markboth{Journal of \LaTeX\ Class Files, Vol. 14, No. 8, August 2015}
{Shell \MakeLowercase{\textit{et al.}}: Bare Demo of IEEEtran.cls for IEEE Journals}
\maketitle

\begin{abstract}
A precise and user-friendly manipulation of image content while preserving image fidelity has always been crucial to the field of image editing. Thanks to the power of generative models, recent point-based image editing methods allow users to interactively change the image content with high generalizability by clicking several control points. But the above mentioned editing process is usually based on the assumption that features stay constant in the motion supervision step from initial to target points. In this work, we conduct a comprehensive investigation in the feature space of diffusion models, and find that features change acutely under in-plane rotation. Based on this, we propose a novel approach named RotationDrag, which significantly improves point-based image editing performance when users intend to in-plane rotate the image content. Our method tracks handle points more precisely by utilizing the feature map of the rotated images, thus ensuring precise optimization and high image fidelity. Furthermore, we build a in-plane rotation focused benchmark called RotateBench, the first benchmark to evaluate the performance of point-based image editing method under in-plane rotation scenario on both real images and generated images. A thorough user study demonstrates the superior capability in accomplishing in-plane rotation that users intend to achieve, comparing the DragDiffusion baseline and other existing diffusion-based methods. See the project page https://github.com/Tony-Lowe/RotationDrag for code and experiment results.
\end{abstract}


\IEEEpeerreviewmaketitle

\section{Introduction}

\IEEEPARstart{I}{mage} editing using generative models has gained great attention and achieved splendid results\cite{roich2022pivotal,kawar2023imagic,brooks2023instructpix2pix,yang2023paint}. It is crucial for personalized image manipulation, {\em e.g.} changing pose, and interactive image creation. Recently, Pan {\em et al.}\cite{pan2023drag} has proposed DragGAN, a point-based Image editing method. DragGAN manipulates the generated image by optimizing latent code of StyleGAN\cite{karras2019style}. To be more specific, DragGAN allows users to provide pairs of handle and target points on the image and optimizes the latent code around the handle points towards target points one pixel-step at a time. Afterward, new handle points are localized by searching the features of initial ones in the updated feature map.

\begin{figure}[!ht]
    \centering
    \subfigure[Original Image]{
    \label{Fig:fig1input}
    \includegraphics[width=.3\columnwidth]{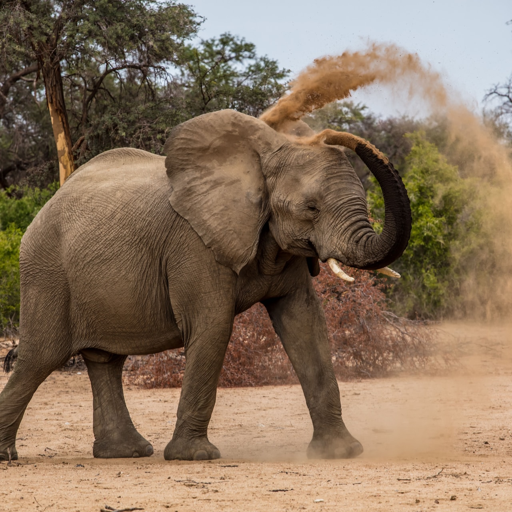}
    }
    \hspace{-0.04\columnwidth}
    \subfigure[User Edit]{
    \label{Fig:userinput}
    \includegraphics[width=0.3\columnwidth]{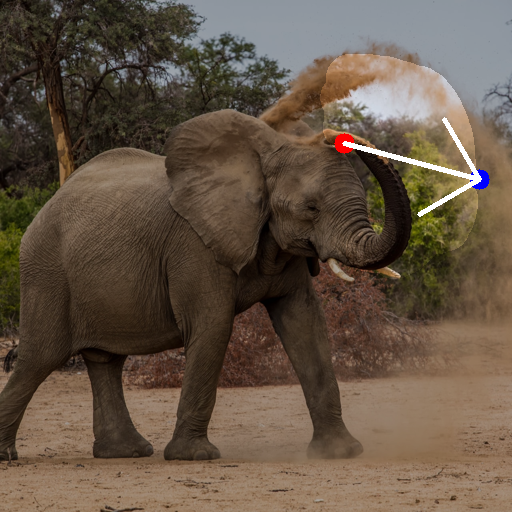}
    }
    \hspace{-0.04\columnwidth}
    \subfigure[DragDiff]{
    \label{Fig:fig1dragdiffusion}
    \includegraphics[width=0.3\columnwidth]{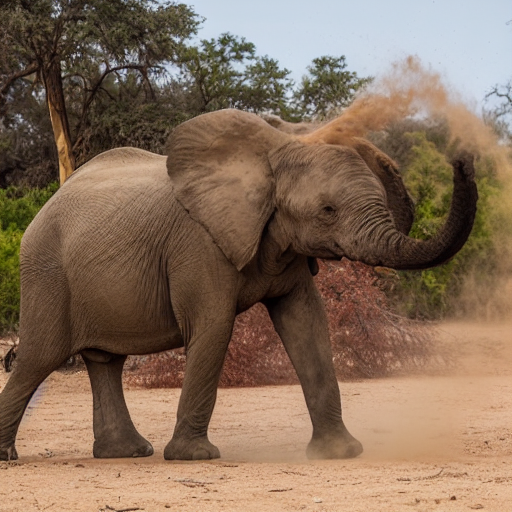}
    }  \\
    \subfigure[SDE-Drag]{
    \label{Fig:fig1SDE}
    \includegraphics[width=0.3\columnwidth]{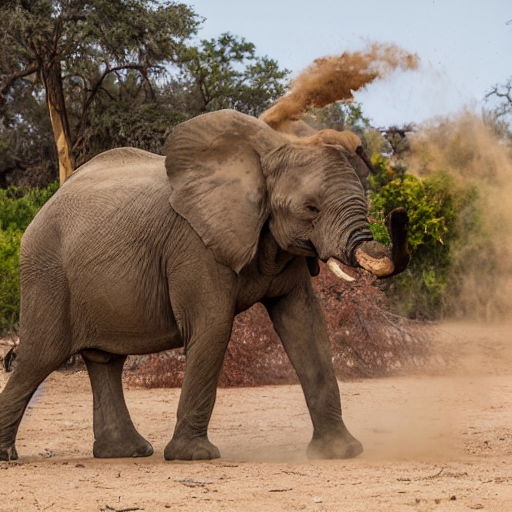}
    }
    \hspace{-0.04\columnwidth}
    \subfigure[FreeDrag]{
    \label{Fig:fig1free}
    \includegraphics[width=0.3\columnwidth]{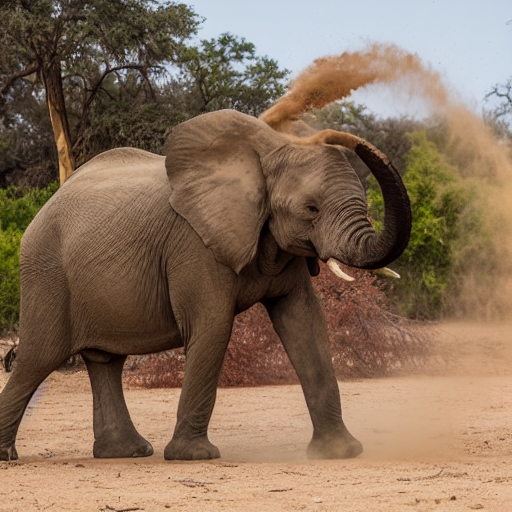}
    }
    \hspace{-0.04\columnwidth}
    \subfigure[Ours]{
    \label{Fig:fig1Ours}
    \includegraphics[width=0.3\columnwidth]{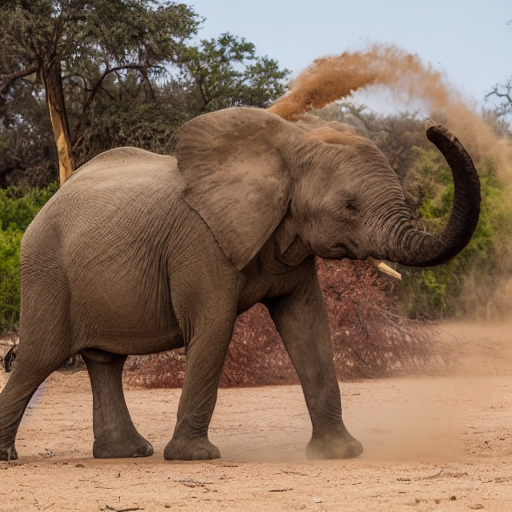}
    }
    
    \caption{Rotation Drag \textbf{significantly improves the point-based editing performance under rotation scenario}. Given an input image, user provide pairs of handle points(red), target points(blue) and mask determining the editing region.}
    \label{fig:Introduction}
\end{figure}

Although DragGAN has successfully achieved pixel-level image manipulation, it inherits the limitation in generality of generative adversarial networks (GANs)\cite{goodfellow2020generative}. Numerous attempts have been made to overcome the limitation of DragGAN. Recent diffusion models\cite{sohl2015deep,ho2020denoising,rombach2022high,saharia2022photorealistic} have shown superior capacity over GANs in the area of image generation. Shi {\em et al.}\cite{shi2023dragdiffusion} propose DragDiffusion, which replaces GANs with diffusion models for point-based image editing. DragDiffusion first fine-tunes the LoRA\cite{hu2021lora} to retain image fidelity during editing. Then the latent codes of handle points are optimized towards the target points, using the feature map generated by UNet\cite{ronneberger2015u} to supervise the motion changes. Nie {\em et al.}\cite{nie2023blessing} propose SDE-Drag that employs diffusion models based on stochastic differential equation (SDE) instead of commonly used probability flow ordinary differential equation (ODE). In terms of editing, SDE-Drag raises a copy-and-paste method, which eliminates optimization process and copies the latent code of the handle points directly towards the target points. Intuitively, both DragDiffusion and SDE-Drag outperform DragGAN thanks to the superior ability of diffusion models. Nevertheless, all the above mentioned methods make an assumption that the feature of the handle points remains unchanged throughout the editing process. When preforming point tracking, they locate the new handle point solely dependent on the feature of the initial handle points. This leads to losing track of the handle points or tracking the wrong regions resemble to the feature of the initial handle points during optimization, resulting in inaccurate or undesired editing results. To tackle this problem, Ling {\em et al.}\cite{ling2023freedrag} propose FreeDrag based on GANs. FreeDrag replaces the point tracking with fuzzy localization. During fuzzy localization, template feature is put forward to preserve the weighted feature of each step. Although FreeDrag outperforms DragGAN, the template feature itself lacks interpretability and it cannot guarantee successful localization when the feature patch changes significantly during the dragging procedure. Take dragging a content over the brick wall towards the sky for example. When intersection is reached, the template feature will gradually lose the feature of the dragging content after numerous weighted sum, leading to optimizing incorrect handle points. 


A natural question arises: under what condition does feature space change acutely? To answer the question and improve the point tracking accuracy of the point-based method using diffusion models, we first investigate the feature space of the UNet. Tang {\em et al.}\cite{tang2023emergent} find that the diffusion models possess rich semantic and geometry information in the feature map of UNet decoder layers. Following their work, we take a step further to examine how diffusion models perform under affine transformations like translation, rotation, scaling, etc and find that the diffusion model performs poorly in the in-plane rotation scenario. The feature map changes drastically when the input image is rotated. Meanwhile we observe that in real world editing, a large portion of the dragging falls into the rotation category. For instance, users might want to wave the tail of animals or make the leaning tower up-straight.

This observation motivates us to use feature map of rotated images to guide the point tracking. To overcome the above mentioned issues, we propose RotationDrag, the first point-based image editing method utilizing feature map of the rotated image to guide the point tracking. In the process of point tracking, we first calculate the rotation angle between the initial and current handle points. Then, we rotate the input image according to the rotation angle to obtain the feature map and use it to locate the new handle points. By utilizing rotated image's feature map, we ensure a more reliable and precise point-based Image editing under rotation scenario, therefore obtaining better optimization results (see Fig. \ref{fig:Introduction}). For evaluation, we present a in-plane rotation focused dataset for point-based editing called Rotatiobench. Both real world photographs and generated images are included to ensure diversity. We attach drag configuration for each image, which includes mask, prompt and coordinates of handle and target points. Moreover, a comprehensive user study on the RotationBench demonstrates the supremacy over the DragDiffusion baseline and other existing diffusion-based methods under in-plane rotation scenario.


To summarize, our key contributions are as follows:
\begin{itemize}
    \item We propose RotationDrag, a simple yet effective point-based image editing that utilizes feature map of rotated image to improve the point tracking accuracy.
    \item We introduce RotationBench, the first benchmark dataset focusing on evaluating point-based editing methods performance under in-plane rotation scenario.  
    \item Experiments demonstrate the advantages of RotationDrag in point-based editing under in-plane rotation scenario. 
\end{itemize}

\begin{figure}
\centerline{\includegraphics[width=\columnwidth]{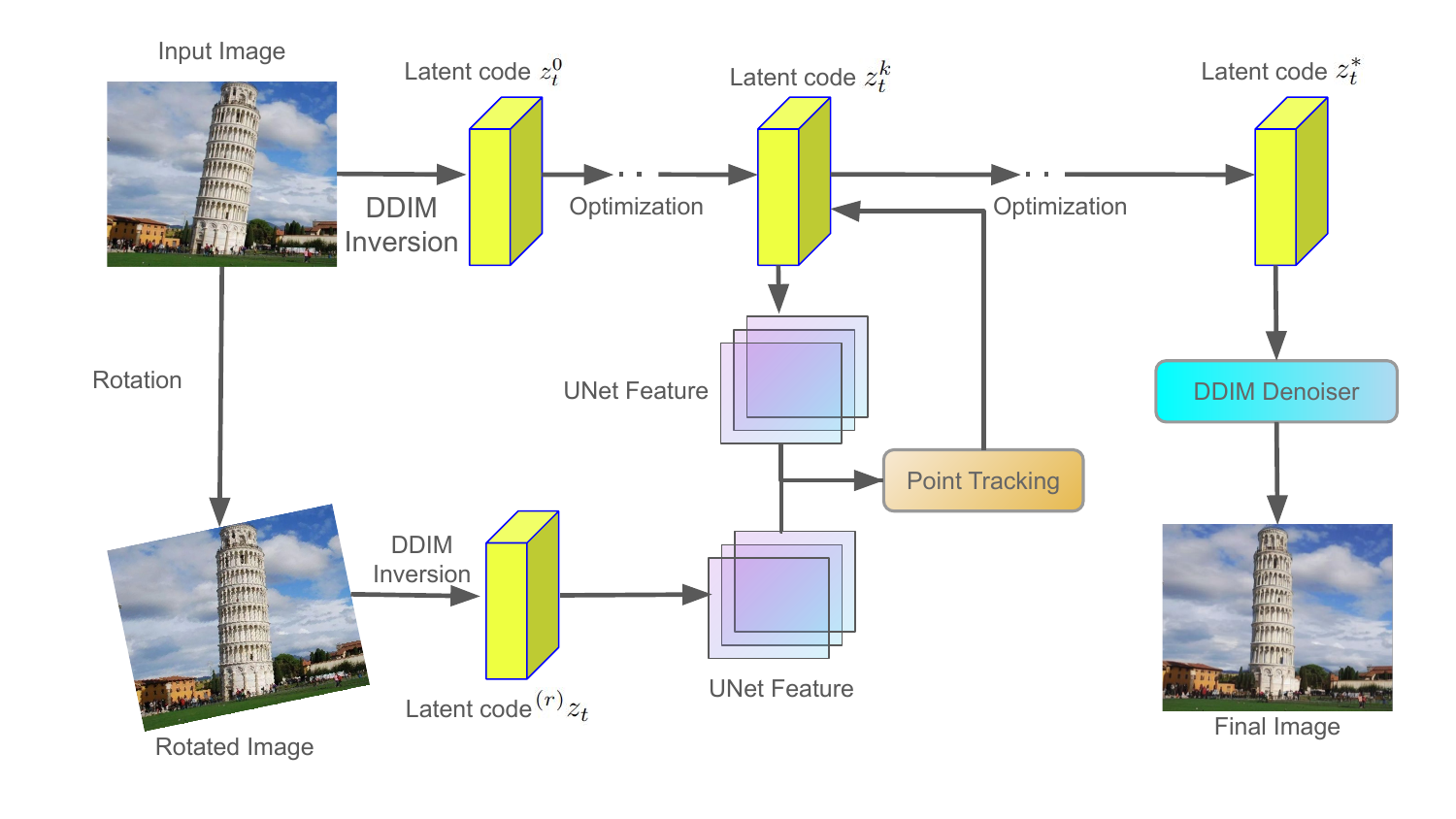}}
\caption{Overview of RotationDrag. Given an input image, we first obtain the latent code of the input image through DDIM\cite{song2020denoising} Inversion. Then we optimize the latent code step-by-step. During optimization, latent code of rotated image is used for UNet feature extraction, ensuring a more reliable point tracking. When optimization is finished, the latent code will go through DDIM Denoiser to restore the edited image. }
\label{fig:overview}
\end{figure}

\section{Method}
\subsection{Preliminaries}

\subsubsection{Diffusion Models}{Denoising diffusion probabilistic models (DDPM)\cite{sohl2015deep,ho2020denoising} are in the category of latent generative models. The idea behind DDPMs is that the data distribution $x$ can be considered as a Markov chain of noisy transformations, where noise $z$ is removed iteratively. The objective is to learn a denoising function that can reverse the noise process and recover the original data from a random noise input. DDPMs define a diffusion process as follows:
\begin{equation}
    x_t = \sqrt{\alpha_t}x_{0} + \sqrt{1-\alpha_t}z_{t},
\end{equation}
where $z_t \sim \mathcal{N}(0,I)$ is the random sampled noise, $\alpha_t$ is the noise level at step $t$ and $T$ is the total number of timesteps. The denoising function is then a conditional distribution $q_{\theta}(x_{t-1}|x_t)$ which is parameterized by the neural network $\theta$. Once trained, $\theta$ can reverse the diffusion process to restore the image data from noisy data. $\theta$ is usually implemented as UNet\cite{ronneberger2015u} in image generation. }

\subsubsection{Point-Based Image Editing}{Point-based image editing methods utilizes feature space of generative models to perform point tracking and motion supervision. Specifically, DragDiffusion employ feature extracted from the $3^{\text{rd}}$ upsampling block of UNet in diffusion models to supervise the motion changes during editing. }



\subsection{Investigation on UNet Feature map} \label{Investigation}

Tang {\em et al.}\cite{tang2023emergent} evaluate the UNet feature map of SD, namely DIFT, for homography estimation using HPatches benchmark\cite{balntas2017hpatches}. Specifically, the feature map extracted from the $3^{\text{rd}}$ upsampling block of the UNet has the strongest geometric feature representation, which is deployed to track handle points in DragDiffusion. 
\begin{table}[!ht]
    \centering
    \caption{Homography estimation accuracy [\%] at 3 pixels. The Feature space of diffusion's UNet changes drastically under in-plane rotaion scenario. }
    \label{tab:homograph}
    \begin{tabular}{lcccc}
    \hline
        Method & Scaling & Rotation & Perspective & Translation \\ \hline
        SIFT\cite{lowe2004distinctive}  & 99.6 & 100.0 & 99.6 & 99.298 \\  
        Superpoint\cite{detone2018superpoint}  & 97.2 & 26.0 & 98.9 & 98.2 \\
        DIFT\cite{tang2023emergent}  & 95.4 & \textcolor{red}{\textbf{37.5}} & 98.6 & 97.9 \\ \hline
    \end{tabular}
\end{table}

We further investigate the performance of SD on affine transformations with specifically curated datasets. Using a subset of HPatches, we crop the reference image to ensure the whole transformation is performed within the image boundaries, warp the image using exactly one type of affine transformation and categorize them accordingly. The warped images are classified into four distinct categories: Scaling, Rotation, Perspective, and Translation.


Following Superpoint\cite{detone2018superpoint}, we use the corner correctness metric for evaluation. We report the comparison of homography accuracy in Table \ref{tab:homograph}. As vividly shown, DIFT exhibits subpar performance in the in-plane rotation scenario. Our conjecture underlying this observation is that the training objective of SD primarily focuses on generative tasks. The training dataset lacks in-plane rotated images. Consequently, SD encounters challenges in accurately interpreting in-plane rotated images due to this deficiency in its training data. 

\subsection{RotationDrag}

Considering subpar performance of SD in the rotational transformations and the majority of the dragging can be categorized into in-plane rotation in real world editing, we propose a in-plane rotation focused approach named RotationDrag to improve the performance of diffusion models in point-based image editing, as shown in Fig. \ref{fig:overview}.

We first process the the user input, from which we can obtain the original image $I$, source points $ \{s_i = (x_{s,i}, y_{s,i})|i=1,2,...,n\} $, target points $ \{t_i = (x_{t,i}, y_{t,i})|i=1,2,...,n\} $ and the binary mask $M$. We consider the point as the rotational axis if source point and handle point overlap, denoted as $\{c_i = (x_{c,i}, y_{c,i})|i=1,2,...,n\}$. If overlapped points do not exist, we position the rotational axis at the furthest extremity of the line perpendicular to the dragging direction within the mask. Then, we invert the image into the latent code $z_t$ after $t$ steps of DDIM inversion\cite{song2020denoising} and optimize it towards the target point. Each optimization step can be divided into 2 parts: motion supervision and point tracking. We will illustrate them in detail in the following part of this section. When optimization is done, the processed latent code $z_t^*$ is fed into the DDIM denoiser to obtain the final results in image.

\subsubsection{Motion Supervision}{Take the handle points and optimized latent code at the $k$-th step optimization as $ \{h^k_i=(x_{h,i}^k, y_{h,i}^k)|i=1,2,...,n\} $ and $z^k_t$. The square patch around $h^k_i$ is defined as $\Omega(h^k_i,r_1)=\{(x,y)|\lvert x-x_{h,i}^k \rvert \leq r_1, \lvert y-y_{h,i}^k \rvert \leq r_1 \}$. The goal of the optimization is to optimize the patch around handle points towards the target point, while keeping the content outside the mask intact. We supervise the optimization using the latent feature map $F_{z_t^k}$ with strong geometric representation extracted from UNet. Same as DragDiffusion\cite{shi2023dragdiffusion}, the motion supervision can be formulated as follows:
\begin{equation}
\begin{aligned}
    L_{motion}= \sum^n_{i=0}\sum_{q_i \in \Omega(h^k_i,r_1)} \lVert F(z_t^k, q_i)-F(z_t^k, q_i+d_i) \rVert_1 \\
    + \lambda \lVert (z_{t-1}^k-z_{t-1}) (1-M) \rVert_1,
\end{aligned}
\end{equation}
where $F(z_t^k,q_i)$ denotes the feature values of latent $z_t^k$ at pixel $q_i$, $d_i = \frac{t_i-h^k_i}{\lVert t_i - h^k_i \rVert_2 }$ is the normalized directional vector pointing from $h^k_i$ towards $t_i$. }

\begin{figure*}[!ht]
    \centering
    \subfigure[Input Image]{
    \begin{minipage}[b]{0.3\columnwidth}
        \includegraphics[width=\columnwidth]{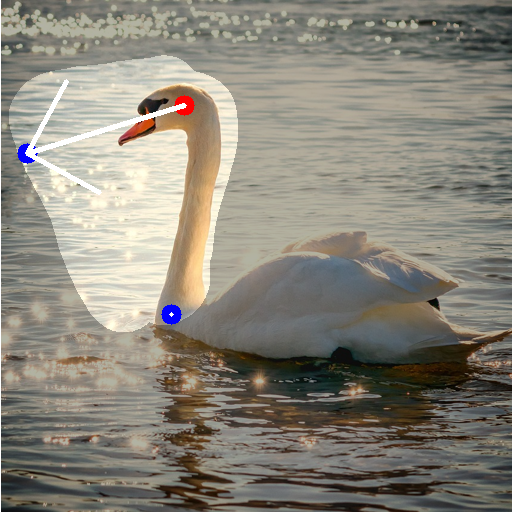} \\
        \includegraphics[width=\columnwidth]{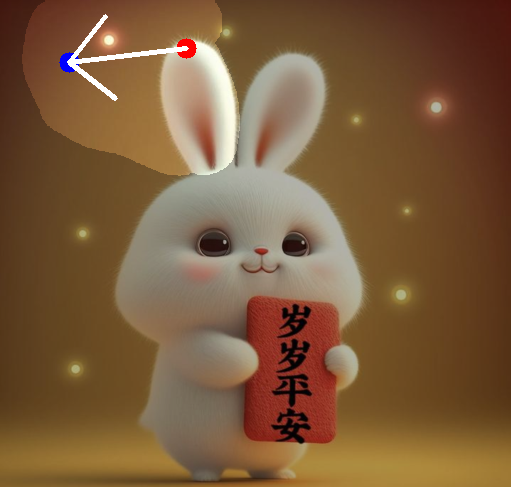} \\
        \includegraphics[width=\columnwidth]{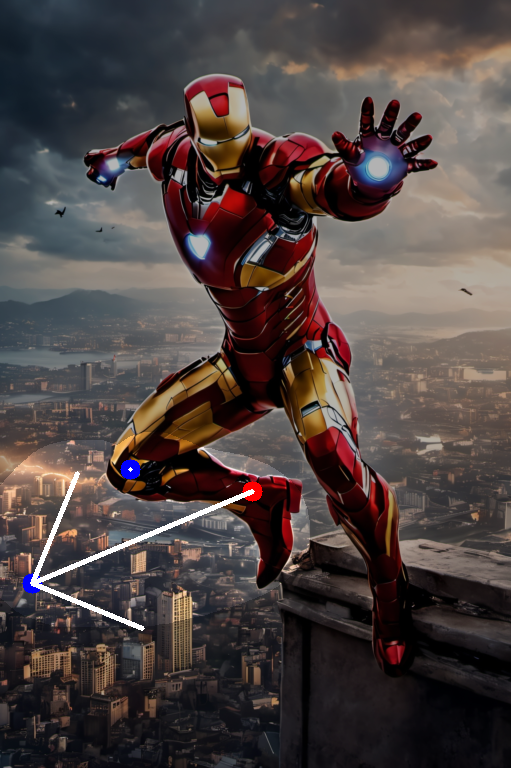}
        
    \end{minipage}
    }
    \hspace{-4.5mm}
    \subfigure[Dragdiff]{
    \begin{minipage}[b]{0.3\columnwidth}
        \includegraphics[width=\columnwidth]{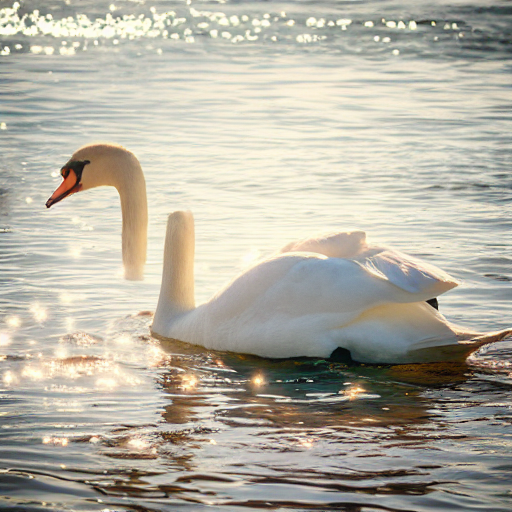} \\
        \includegraphics[width=\columnwidth]{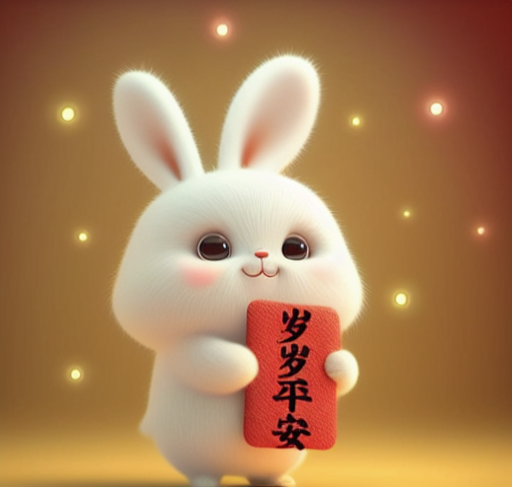} \\
        \includegraphics[width=\columnwidth]{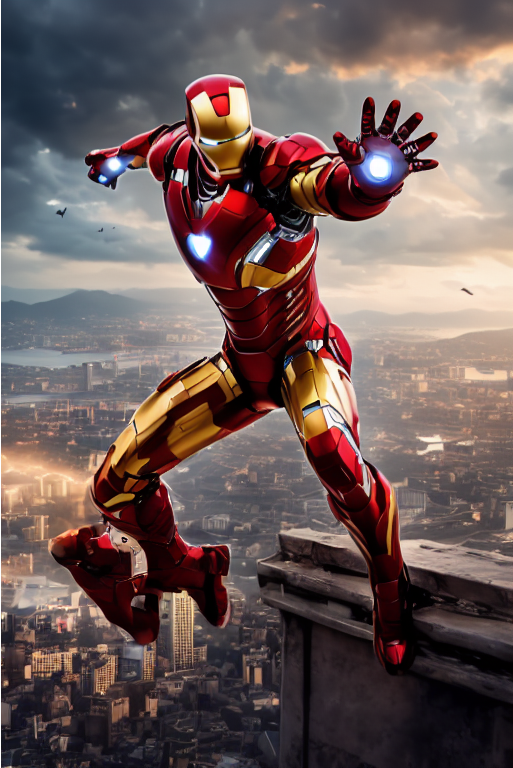}
    \end{minipage}
    }
    \hspace{-4.5mm}
    \subfigure[FreeDrag]{
    \begin{minipage}[b]{0.3\columnwidth}
        \includegraphics[width=\columnwidth]{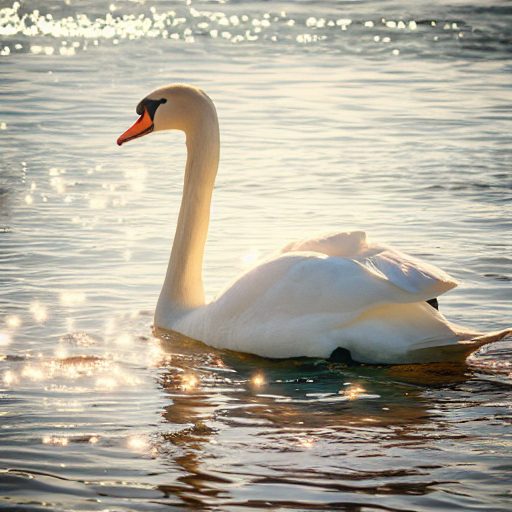} \\
        \includegraphics[width=\columnwidth]{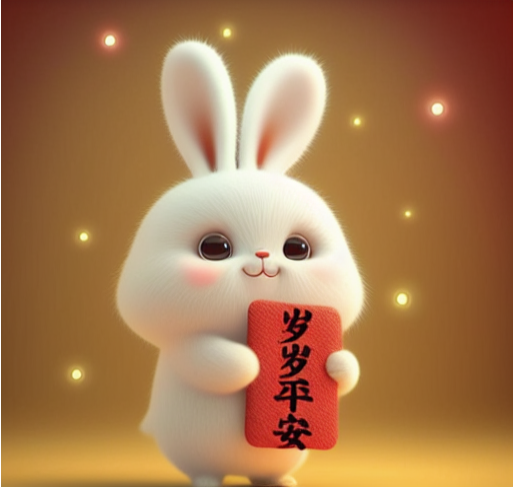} \\
        \includegraphics[width=\columnwidth]{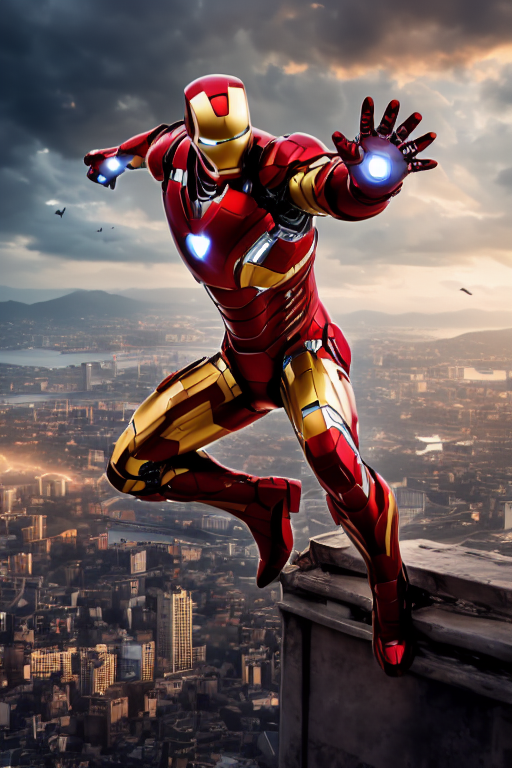}
    \end{minipage}
    }
    \hspace{-4.5mm}
    \subfigure[SDE-Drag]{
    \begin{minipage}[b]{0.3\columnwidth}
        \includegraphics[width=\columnwidth]{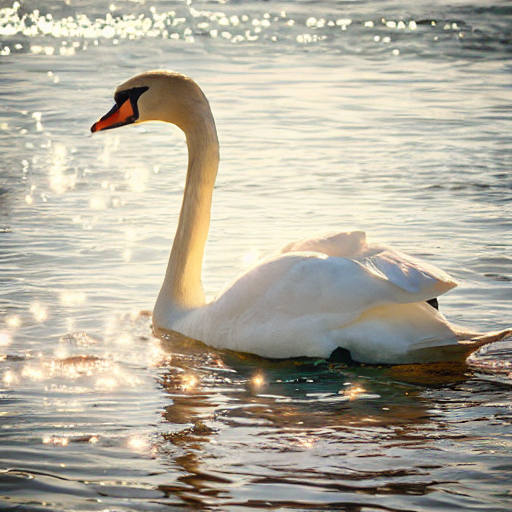} \\
        \includegraphics[width=\columnwidth]{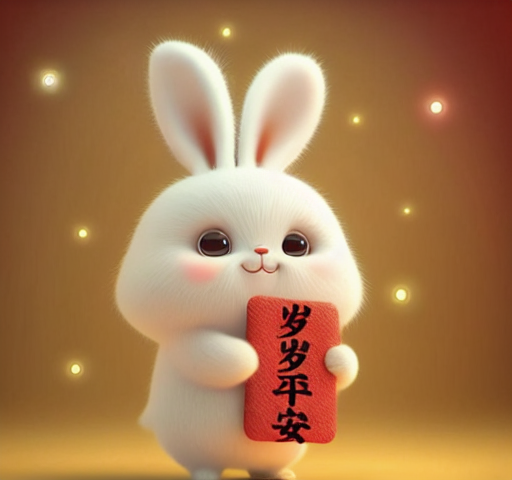} \\
        \includegraphics[width=\columnwidth]{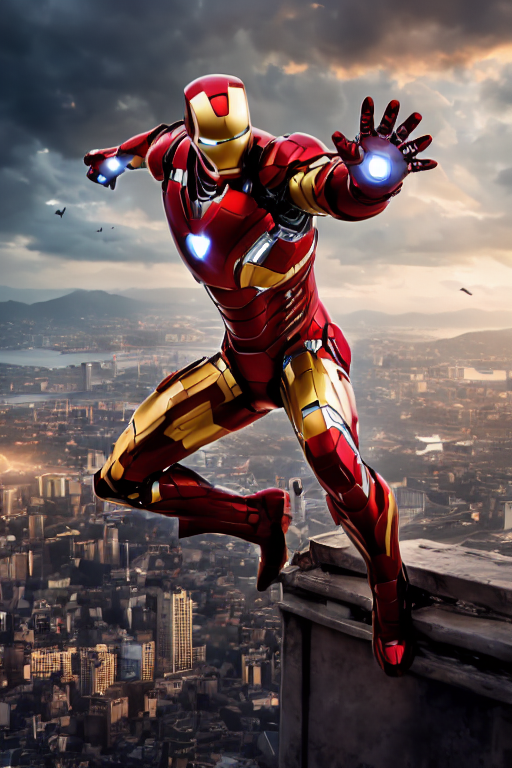}
    \end{minipage}
    }
    \hspace{-4.5mm}
    \subfigure[Ours]{
    \begin{minipage}[b]{0.3\columnwidth}
        \includegraphics[width=\columnwidth]{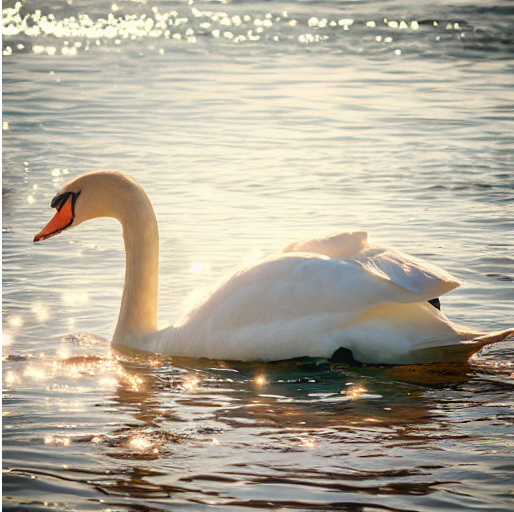} \\
        \includegraphics[width=\columnwidth]{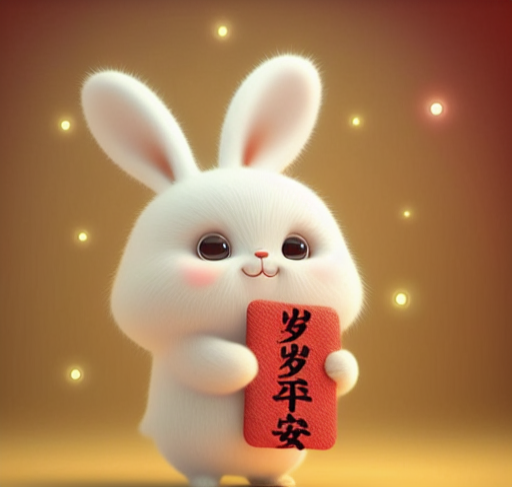} \\
        \includegraphics[width=\columnwidth]{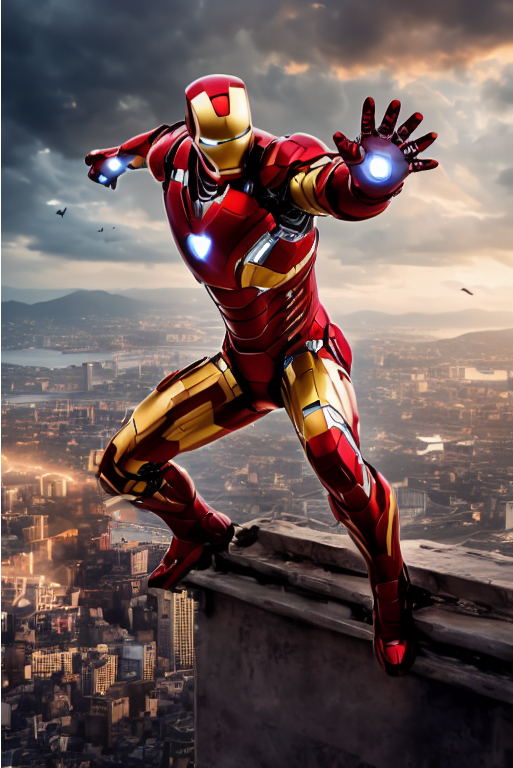}
    \end{minipage}
    }
    \caption{Visual comparison between DragDiffusion, FreeDrag(our diffusion implementation version), SDE-Drag and RotationDrag. The left column displays the input images, while columns in the right displays editing results of DragDiffusion, Diffuison version FreeDrag of our implementation, FreeDrag, SDE-Drag and RotationDrag respectively.}
    \label{fig:Qualitative}
\end{figure*}

\subsubsection{Point Tracking}{To ensure the dragging is performed on the desired direction, the location of the handle points needs to be updated together with the latent as well. After each optimization on latent, we need to relocate the handle points. Dragdiffusion\cite{shi2023dragdiffusion} used the UNet feature maps of the updated latent code and the original latent code to track new handle points. As explained in Sec. \ref{Investigation}, UNet features change immensely under in-plane rotational change, resulting in finding the incorrect handle points. To tackle this problem, we need to recognize UNet feature representation of the latent around the handle points under the current rotation angle. The simplest way is to rotate the input image accordingly to generate feature map for more precise point tracking. We conduct another DDIM inversion to obtain the latent of the rotated image and used it to extract feature vectors of the rotated source points. The angle of the rotation for handle point $h^k_i$ is computed as follows:
\begin{equation}
    \theta_i^k = \arctan(\frac{y_{h,i}-y_{c,i}}{x_{h,i}-x_{c,i}}) - \arctan(\frac{y_{s,i}-y_{c,i}}{x_{s,i}-x_{c,i}}).
\end{equation}
The final result of image rotation is independent of the rotation axis, and even if the rotated source point does not coincide with the handle point, it is permissible to perform rotation solely based on the angle due to minor effects translation had on UNet feature. The rotated image and the corresponding source point is denoted as $^{(r)}I$, $^{(r)}s_i$. The $t$-th inversion step latent of the rotated image is denoted as $^{(r)}z_t$. We use the UNet feature maps $F(^{(r)}z_t)$ and $F(z_t^{k+1})$ to track the new handle points. To update the handle points, a nearest neighbour search of the handle points is performed as follows:
\begin{equation}
    h_i^{k+1} = \arg\min_{q \in \Omega(h_i^k, r_2)} \lVert F(z_t^{k+1}, q)-F(^{(r)}z_t, ^{(r)}s_i) \rVert.
\end{equation}
}

\section{Experiments}
\subsection{Implementation Details and Dataset}

Following DragDiffusion\cite{shi2023dragdiffusion}, We use Stable Diffusion 1.5\cite{rombach2022high} as our diffusion model and finetune LoRA\cite{hu2021lora} before the optimization. During optimization, we adopt Adam optimizer\cite{kingma2014adam} with a learning rate of 0.01 to optimize the latent code. The hyper parameters are set to be $r_1=1,r_2=3,\lambda=0.1$. In our implementation, the optimization stops when the distance between the handle points and the corresponding target points is smaller than 2 pixels. The maximum optimization step is 160. 

Since point-based image editing is a relatively new method and few people focused on in-plane rotation, the existing benchmark cannot illustrate the performance of the editing method under in-plane rotation properly. Therefore we present RotationBench, a dataset built upon both real world and generated images focusing on in-plane rotation. We provide drag configuration file for every image, including binary mask, prompt and points coordinates for dragging. 


\subsection{Experimental Results}

\subsubsection{Qualitative Results}{Fig. \ref{fig:Qualitative} shows the qualitative comparsion between DragDiffusion, FreeDrag(our diffusion implementation version), SDE-Drag and our method. As shown in the figure, our RotationDrag accurately moves the handle points towards target points and achieves highest quality on both real-world ($1^{\text{st}}$ row of Fig. \ref{fig:Qualitative}), art ($2^{\text{nd}}$ row of Fig. \ref{fig:Qualitative}) and generated images ($3^{\text{rd}}$ row of Fig. \ref{fig:Qualitative}). Specifically, when rotational axis is not given in drag instructions, our method can not only produce content closer to the destination but also maintain reasonable image content as shown in the $2^{\text{nd}}$ row of Fig. \ref{fig:Qualitative}. These results demonstrate the superior ability of RotationDrag to maintain the image fidelity while performing intense dragging or rotating.} 

\begin{figure}
    \centering
    \includegraphics[width=0.6\columnwidth]{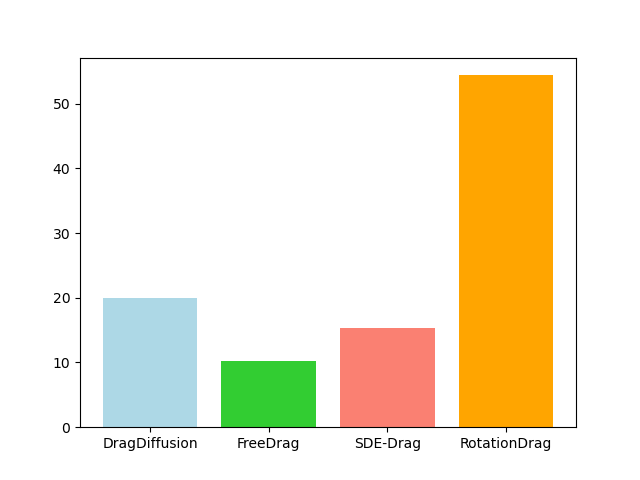}
    \caption{Results of user study. RotationDrag outperforms all competitors by a large margin.}
    \label{fig:bar}
\end{figure}

\subsubsection{User Study}{There were 20 participants and 25 questions for each comparison between DragDiffusion, FreeDrag, SDE-Drag and our method by default. In each question, participants were presented with the original image, the user editing including handle points, target points and masks on the original image and 4 edited images produced by distinct models. The participants were asked to select the best editing result in the 4 edited images. As shown in Fig. \ref{fig:bar}, RotationDrag outperforms all competitors by a large margin.}

\subsubsection{Discussion}{Even though our method has achieved remarkable results under in-plane rotation scenario, it is still slower than original DragDiffusion. The reason behind this phenomenon is that inversion is performed in each point tracking procedure. Weak performance in rotation of the Stable Diffusion network is the main cause for this approach. We did try to rotate the latent code to produce a rotated image, only to produce distorted images. We will explore the possibility of teaching stable diffusion for a better knowledge of rotation.} 

\section{Conclusion}

In this work, we investigate the UNet features map's performance on affine transformation and report the subpar performance under in-plane rotation scenario. Based on this insight, we propose RotationDrag to boost the performance of Stable Diffusion under in-plane rotation scenario in interactive point-based image editing. By utilizing feature map of rotated images, we track better point movement and achieve high quality dragging results. Experiments are conducted to demonstrate the superiority of RotationDrag when dealing in-plane rotation. 



\bibliographystyle{IEEEtran}

\end{document}